\documentclass[letterpaper, 10 pt, conference]{ieeeconf}  

\IEEEoverridecommandlockouts                              

\usepackage{hyperref}
\usepackage[noadjust]{cite}

\usepackage{graphicx}
\let\labelindent\relax
\usepackage{enumitem}
\usepackage{bm}
\usepackage{bbm}
\usepackage{amsmath}
\usepackage{amssymb}
\usepackage{gensymb}
\DeclareMathOperator*{\argmin}{\arg\!\min}

\usepackage{subcaption}
\usepackage{adjustbox}
\usepackage{multirow}

\newcommand\blfootnote[1]{%
  \begingroup
  \renewcommand\thefootnote{}\footnote{#1}%
  \addtocounter{footnote}{-1}%
  \endgroup
}

\newcommand\Tstrut{\rule{0pt}{2.6ex}}         
\newcommand\Bstrut{\rule[-0.9ex]{0pt}{0pt}}   

\usepackage{soul}
\usepackage[dvipsnames]{xcolor}

\DeclareMathOperator*{\expect}{{\huge \mathbb{E}}}
\newcommand{\cbar}{\, | \,}

\title{\LARGE \bf
Residual Reinforcement Learning from Demonstrations}

\author{Minttu Alakuijala$^{1,2,3}$, Gabriel Dulac-Arnold$^{1}$, Julien Mairal$^{2}$, Jean Ponce$^{3}$ and Cordelia Schmid$^{1}$
\thanks{
$^{1}$Google Research
}
\thanks{
{\tt\small \{minttu,dulacarnold,cordelias\}@google.com}
}%
\thanks{
$^{2}$Inria, Univ. Grenoble Alpes, CNRS, Grenoble INP, LJK
}%
\thanks{
$^{3}$Inria, \'{E}cole Normale Sup\'{e}rieure, ENS, CNRS, PSL Research University
}%
\thanks{
\tt\small \{julien.mairal,jean.ponce\}@inria.fr}
}

\begin{document}

\maketitle
\thispagestyle{empty}
\pagestyle{empty}

\begin{abstract}
Residual reinforcement learning (RL) has been proposed as a way to solve challenging robotic tasks by adapting control actions from a conventional feedback controller to maximize a reward signal. We extend the residual formulation to learn from visual inputs and sparse rewards using demonstrations. Learning from images, proprioceptive inputs and a sparse task-completion reward relaxes the requirement of accessing full state features, such as object and target positions. In addition, replacing the base controller with a policy learned from demonstrations removes the dependency on a hand-engineered controller in favour of a dataset of demonstrations, which can be provided by non-experts. Our experimental evaluation on simulated manipulation tasks on a 6-DoF UR5 arm and a 28-DoF dexterous hand demonstrates that residual RL from demonstrations is able to generalize to unseen environment conditions more flexibly than either behavioral cloning or RL fine-tuning, and is capable of solving high-dimensional, sparse-reward tasks out of reach for RL from scratch. \blfootnote{\newline \indent Videos of policy executions are available on the project website:\\ \indent \url{https://sites.google.com/view/rrlfd/}.}

\end{abstract}

\section{INTRODUCTION}

Reinforcement learning has the potential to enable autonomous learning of hard-to-specify behaviors under varying environment conditions. However, learning control entirely from data requires a significant amount of robot experience to be collected for most tasks of interest. To improve sample efficiency, residual RL~\cite{johannink2019residual,silver2018residual} takes advantage of a conventional controller to address the part of the task that can be solved efficiently by feedback control, but adapts the control outputs through a superimposed residual policy, trained with RL.

\looseness=-1
However, relying on hand-engineered base controllers imposes certain limitations on the kind of tasks that can be addressed.
Providing prior information in the form of demonstrations instead of a manually specified feedback controller would allow residual RL to be used in a wider set of applications for which first-principles physical modeling and accurate state estimation are not feasible.
Drawing from the extensive field of imitation learning, we train residual RL to complement a policy learned with behavioral cloning~(BC)~\cite{pomerleau1989alvinn}.

Replacing the base controller with a policy learned from demonstrations presents a number of advantages. First, the resulting policy can be trained from data alone, in the form of demonstrations for the base controller and rollouts in the environment for the residual policy. We argue that in many settings, demonstrations are easier to provide than custom controllers for new tasks, objects or robot configurations. Demonstrations may, for example, be obtained by teleoperation, or any logged task execution data. Second, feedback controllers typically rely on full state estimation, such as the positions of objects, targets and obstacles. We instead propose to learn control entirely from visual and robot proprioceptive inputs, and learn task-specific state features through BC on demonstrated trajectories. Whereas the residual policies considered in prior work require hand-crafted state features~\cite{johannink2019residual,silver2018residual,barekatain2019multipolar,schoettler2019deep,rana2019residual,davchev2020residual}, our method, Residual Reinforcement Learning from Demonstrations (RRLfD), 
learns from images and requires minimal feature engineering.

Moreover, learning a residual on top of a base controller enables a significant decrease in sample complexity compared to training RL from scratch~\cite{johannink2019residual,silver2018residual,schoettler2019deep,rana2019residual,davchev2020residual} as the controller provides a good prior for the RL policy. This reduces the importance of exploration and thus of reward shaping, and allows for more difficult and longer-horizon tasks to be learned from sparse rewards compared to a randomly exploring agent that might never observe a task completion in practice. The residual formulation also simplifies the learning task, allowing the RL policy to take smaller actions and to reduce the magnitude of exploration required---an important consideration for robotics applications. Moreover, the residual form has potential to alleviate catastrophic forgetting~\cite{mccloskey1989catastrophic} compared to a policy that is simply initialized from demonstrations and fine-tuned with RL.

Our contribution is therefore twofold: we propose a novel way to leverage data gathered from the system to accelerate the training of residual RL policies in image space through visual feature sharing, and present a fully data-driven method, RRLfD, leveraging both imitation learning and residual RL to allow for efficient training of sparse image-based tasks from demonstrations. We evaluate our method on robotic manipulation tasks; however, the formulation is general and readily applicable to any continuous control task. We begin by comparing the proposed method to prior work in Section \ref{sec:related_work}, then describe the residual policy's structure and training in Section \ref{sec:residual_policy}. We evaluate our method in seven experimental settings from two simulated task suites and present the results in Section \ref{sec:experiments}.

\begin{figure*}[t]
\centering
\begin{subfigure}{0.45\textwidth}
\includegraphics[height=5.53cm]{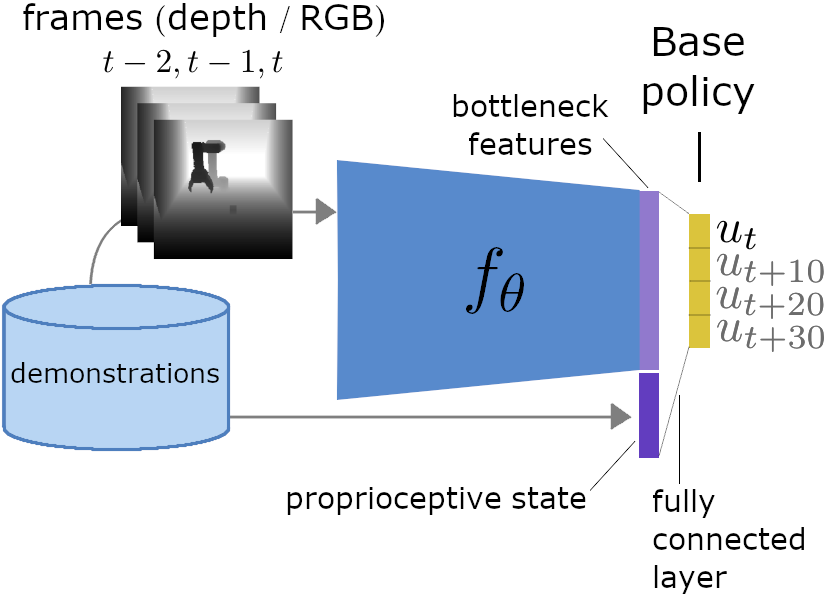}
\subcaption{}
\label{fig:overview_left}
\end{subfigure}
\begin{subfigure}{0.54\textwidth}
\includegraphics[height=5.53cm]{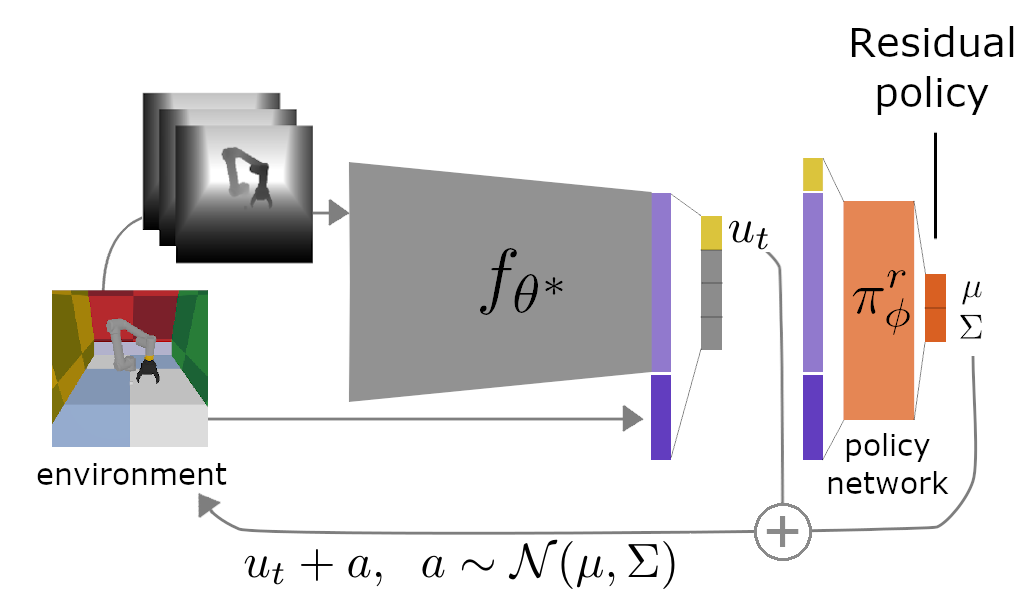}
\subcaption{}
\label{fig:overview_right}
\end{subfigure}
\caption{a) We propose a way to leverage demonstration data to learn a control policy as well as task-relevant visual features through behavioral cloning on image and proprioceptive inputs. b) The policy is then improved through reinforcement learning by a superimposed residual policy, based on the learned visual features, allowing data-efficient learning of control policies in image space from sparse rewards.}
\label{fig:overview}
\end{figure*}

\section{Related work}
\label{sec:related_work}
Residual RL for robot control, concurrently proposed by Johannink \emph{et al.}
\cite{johannink2019residual} and Silver \emph{et al.} \cite{silver2018residual},
has been studied for manipulation tasks \cite{johannink2019residual,silver2018residual,schoettler2019deep} and navigation \cite{rana2019residual} to complement a conventional feedback controller. Most related to our method is the work of Davchev \emph{et al.} \cite{davchev2020residual}, where the base policy is structured as a dynamic movement primitive whose coefficients are learned using demonstrations but which operates on full state features, including a goal position for insertion. We instead make no assumptions about the structure of the base policy but train a convolutional neural network using direct behavioural cloning in image space, without requiring access to accurate state estimation.

Another line of work has generalized the residual formulation to consider adaptively re-weighting multiple base policies, each trained with RL to solve the same task under different robot dynamics~\cite{barekatain2019multipolar}. Related to residual control is also a recent work on robotic throwing~\cite{zeng2019tossingbot}, where a scalar release velocity given by
projectile physics is additively corrected using self-supervised learning.
In addition to policies, residual formulations have also been proposed for correcting the predictions of approximate environment models \cite{ajay2018augmenting,allevato2019tunenet,saveriano2017data} and adapting features in transfer learning \cite{zhang2019side}.

Alternatively, an agent's policy could be initialized using demonstrations and its training continued using RL~\cite{kober2009policy,rajeswaran2018learning}.
We instead fix the base policy after training and always add it to the continuous control action in order to alleviate the risk of catastrophic forgetting for parts of the state space.

Also related to our work are off-policy RL methods
which learn from demonstrated data and new environment rollouts in conjunction, by merging them in a single replay buffer~\cite{hester2018deep,vecerik2017leveraging} or with an auxiliary objective that aims to stay close to the demonstrations while also maximizing observed reward~\cite{nair2018overcoming,rajeswaran2018learning,zhu2018reinforcement}. These approaches either pretrain a policy offline and then continue training on the real system, or learn from scratch using offline data as a prior. There is also work on fully offline RL that can be used to train a policy using only demonstrations without access to the system~\cite{gulcehre2020rl, fu2020d4rl, wang2020critic, argenson2020model}. As our approach is rather orthogonal to existing learning from demonstration methods, it could be readily combined with other training protocols (e.g. a pre-populated replay buffer) and objectives (e.g. staying close to demonstrations). An interesting direction for future work could be to extend fully offline methods to further online training, for example by using offline RL in lieu of BC in our method.

Prior work on picking and insertion from images has leveraged both demonstrations and black-box policies.  \cite{schoettler2019deep} look at plug-insertion tasks using both residual RL and RL with demonstrations from images. Although the task is executed on a real robot, the target is fixed and known by the base controller, and the final control policy is not shown to generalize to unseen target positions, unlike our method. The QT-Opt algorithm \cite{kalashnikov2018qt} learns grasping from images using a parallel set of robots (similar to \cite{levine2018learning}) and is initialized with a black-box controller to help with the initial exploration phase. However, it does not use residual control, and the initial controller is hand-coded.  QT-Opt is able to learn a good grasping policy both in simulation and on real robots, but comparing data efficiency is not straightforward as performance is often reported for the number of episodes or gradient updates rather than environment time steps. Our method can instead be initialized from demonstrations without a hand-engineered controller.

\section{Residual Policy Learning from Demonstrations}
\label{sec:residual_policy}
RRLfD is trained in two stages (Fig. \ref{fig:overview}). A convolutional neural network (CNN) is first trained to predict demonstrated control actions given a short history of depth or RGB images as well as current robot proprioceptive state (Section \ref{ssec:base_controller}). Using demonstrated trajectories, the network learns to capture visual features relevant to solving the control task.

However, the base policy can only learn behaviour that was present in the demonstrations. Depending on the coverage of the dataset, behaviors involving recovery from failure or states near the edges of the workspace may not be included. Moreover, pure BC is known to suffer from compounding errors \cite{ross2010efficient}. To improve the policy with autonomous environment interaction, a light-weight policy network on top of the learned CNN features is trained with RL to additively correct the base policy's actions (Section \ref{ssec:residual_controller}).

\subsection{Base policy}
\label{ssec:base_controller}
We learn a base policy using behavioral cloning (BC). First, $N$ demonstrations are gathered for a task of interest to create a dataset consisting of the demonstrated trajectories' states $S^i=[s^i_1,...,s^i_{T_i}]$ and actions $A^i=[a^i_1,...,a^i_{T_i}]$ taken at each time step $t=1,...T_i,i=1,...,N$, drawn from a behavior policy $\pi^e(s_t^i)$. Each state $s$ includes a history of stacked camera frames, as either depth or RGB, as well the robot's proprioceptive state. We consider a history of three previous frames to allow policies to capture velocity and acceleration of objects in the scene. Where applicable, the inverse kinematics of the robot are used to reduce the effective dimensionality of the action space from joint space control to end-effector control.

A convolutional neural network $f_\theta$ mapping a state~$s$ to a sequence of actions $[\hat{a}_t, \hat{a}_{t+10}, \hat{a}_{t+20}, \hat{a}_{t+30}]$, and parametrized by $\theta$, is then trained with BC on this dataset:
\begin{align}
    \theta^* = \argmin_{\theta} \expect_{( s_t,  a_t) \sim (S^{1:N}, A^{1:N})} \sum_{\delta=0, 10, 20, 30} L(\hat{a}_{t+\delta},  a_{t+\delta}),
\end{align}
to obtain a policy $\pi_\theta^b(s)=\hat{a}_t$.
In addition to predicting~$a_t$, we include actions 10, 20, and 30 time steps ahead and minimize loss over each of these targets, as done by \cite{strudel2019learning}. 
This serves as an auxiliary prediction task leading to better data efficiency~\cite{jaderberg2016reinforcement} and encourages the policy to capture longer-term task structure (up to three seconds in the future given control actions at 10Hz).

Depending on the task, the action space may have discrete and continuous components, which we may denote by $g$ and~$v$, respectively, such that $a = \left[v, g\right]$ and $\hat{a} = \left[\hat{v}, \hat{g}\right]$. For tasks requiring discrete components such as gripper state control, the loss $L$ is defined by weighting L2 and cross-entropy terms, as in~\cite{strudel2019learning}:
\begin{equation}
L((\hat{v}, \hat{g}), (v, g)) = \lambda ||\hat{v} - v||_2^2 -(1 - \lambda)\sum_{c=1}^{|g|} g_c \log\hat{g_c},
\end{equation}
where
$\lambda$ is a hyperparameter. For tasks with fully continuous action spaces, only the regression loss is considered ($\lambda=1$).

\subsection{Residual policy}
\label{ssec:residual_controller}
After BC training, the base policy $\pi_\theta^b$ is fixed. Given $u=\pi_\theta^b(s)$, a residual policy $\pi_\phi^r(s, u)=a$ is trained using RL to complement the base action $u$, and the resulting control action $u + a$ is executed in the environment.
$\pi_\phi^r$ is trained to maximize the expected discounted return $G^\pi$ in the residual Markov Decision Process (MDP):
\begin{align}
&Q^\pi(s,a) = \expect \left [ G^\pi(s, a) \right ] = \expect \left [ \sum_{t=0}^T \gamma^t r(s_t, a_t) \right ],\\
& s_t \sim p(\cdot \cbar s_{t-1}, u_{t-1} + a_{t-1}),  a_t \sim \pi_\phi^r(\cdot \cbar s_t, u_t), \nonumber\\
& u_t = \pi_\theta^b(s_t), s_0 = s \nonumber
\end{align}
where $p$ is the state transition dynamics, $\gamma$ is a discount factor in $\left[ 0,1 \right) $ and $r$ is a reward function whose value is $1$ if the task has been completed and $0$ otherwise.
$\pi_\phi^r$ takes as input the base action $u$, the robot's proprioceptive state, and the features of the bottleneck layer (i.e., the final hidden layer before the fully connected output layer) of the BC policy network $f_\theta$. The bottleneck layer features provide the residual policy with visual information learned during BC. $\pi_\phi^r$ then outputs a corrective action $a$ by
predicting the parameters of a Gaussian distribution with diagonal covariance:
\begin{align}
a \sim \mathcal{N}(\mu, \Sigma), \hspace{0.3cm} \text{with } (\mu, \Sigma) = \pi_\phi^r(s, u).
\end{align}
At evaluation time, the policy is made deterministic: $u + \mu$ is executed in the environment instead of drawing a sample from the predicted Gaussian.
	

\begin{figure*}[t]
  \begin{subfigure}{.136\textwidth}
    \centering
    \includegraphics[width=\linewidth]{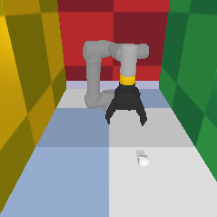}
    \subcaption{Pick}
    \label{fig:pick_task}
  \end{subfigure}
  \begin{subfigure}{.136\textwidth}
    \centering
    \includegraphics[width=\linewidth]{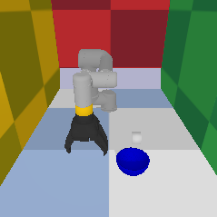}
    \subcaption{Bowl}
    \label{fig:bowl_task}
  \end{subfigure}
  \begin{subfigure}{.136\textwidth}
    \centering
    \includegraphics[width=\linewidth]{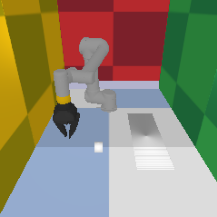}
    \subcaption{Push}
    \label{fig:push_task}
  \end{subfigure}
  \begin{subfigure}{.136\textwidth}
    \centering
    \includegraphics[width=\linewidth]{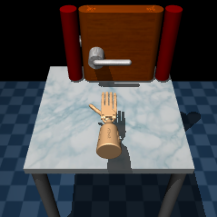}
    \subcaption{Door}
    \label{fig:door_task}
  \end{subfigure}
  \begin{subfigure}{.136\textwidth}
    \centering
    \includegraphics[width=\linewidth]{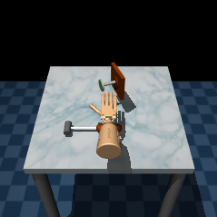}
    \subcaption{Hammer}
    \label{fig:hammer_task}
  \end{subfigure}
  \begin{subfigure}{.136\textwidth}
    \centering
    \includegraphics[width=\linewidth]{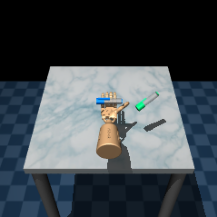}
    \subcaption{Pen}
    \label{fig:pen_task}
  \end{subfigure}
  \begin{subfigure}{.136\textwidth}
    \centering
    \includegraphics[width=\linewidth]{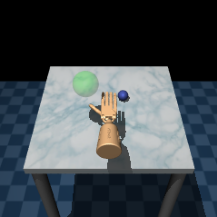}
    \subcaption{Relocate}
    \label{fig:relocate_task}
  \end{subfigure}
  \caption{We evaluate RRLfD on seven manipulation tasks on two different robotic simulation platforms: a 6-DoF UR5 arm (a--c) and a 28-DoF ShadowHand model (d--g).}
\end{figure*}

\begin{figure*}[t]
    \centering
    \includegraphics[width=\linewidth]{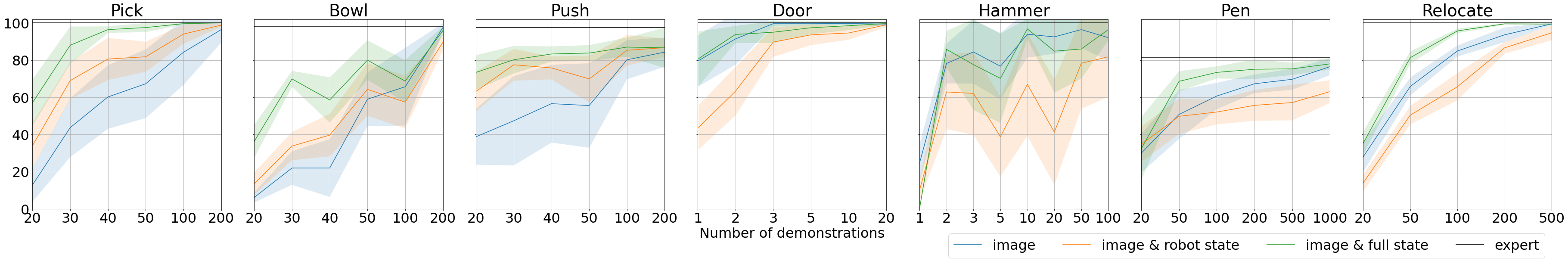}
    \caption{BC success rates (as \%) evaluated on 100 unseen initial states.
    (10 seeds, 95\% confidence intervals)}
    \label{fig:bc_results}
\end{figure*}

\section{Experiments}
\label{sec:experiments}

\subsection{Environments}
To demonstrate the generality of our approach, we consider control tasks from two distinct manipulation suites of different complexity and dimensionality. The Mime environments~\cite{strudel2019learning} define of a 6-DoF UR5 robotic arm with a 3-finger Robotiq gripper using the PyBullet physics simulator~\cite{coumans2019pybullet}. An inverse kinematics submodule allows control in task space by converting end-effector velocities to joint velocities at 10Hz, exposing to the higher-level policies $\pi_\theta^b$ and $\pi_\phi^r$ an action space consisting of the translational velocity $v$ in $\mathbb{R}^3$ and a binary gripper state $g$ in $\{0, 1\}$. We consider two standard tasks included in \cite{strudel2019learning}: picking up a cube (Fig. \ref{fig:pick_task}) as well as picking up a cube and placing it in a bowl (\ref{fig:bowl_task}), and additionally define a third task of pushing a cube to a fixed target location (\ref{fig:push_task}). All object positions and sizes, starting robot pose as well as the positions of the surrounding walls are drawn at random for each episode.

Our second task suite of interest, Adroit \cite{rajeswaran2018learning}, defines a five-fingered arm controlled at 10Hz in a continuous action space consisting of joint values for a ShadowHand-inspired dexterous hand (24 DoF) and the position of the arm (4 DoF).
The suite consists of four MuJoCo \cite{todorov2012mujoco} environments, defining tasks of opening a door, hammering a nail, in-hand orientation of a pen, and object relocation (Fig. \ref{fig:door_task}--\ref{fig:relocate_task}).

\subsection{Demonstrations}
We follow convention set by prior work and define the expert policy $\pi^e$ by either a script (for Mime environments~\cite{strudel2019learning}) or a previously trained RL agent (for Adroit environments \cite{jain2019learning}), but it could equally come from a human teleoperator or any black-box controller.

\subsection{Observation space}
We consider two kinds of inputs to the base policy: images only, and images with robot proprioceptive state. To give an indication of an upper bound reachable with flawless state estimation pipelines, 
we also consider BC trained on images with full state, including both proprioception and the positions of relevant objects. For the Mime environments, the proprioceptive state consists of tool position in $\mathbb{R}^3$, tool velocity in $\mathbb{R}^3$, gripper state in $\mathbb{R}$ and gripper opening (if positive) or closing (if negative) velocity in $\mathbb{R}$. Full state additionally includes cube position in $\mathbb{R}^3$ (and bowl position in $\mathbb{R}^3$, if applicable). For Adroit, proprioception includes joint positions, joint velocities, palm position and the fingertips' tactile sensor readings as defined by \cite{jain2019learning}. As full state, we use the environments' original observation space. We keep each task suite's default setting and use depth cameras in Mime and RGB cameras in Adroit, but use a consistent resolution of $240\times240$px in each.

\subsection{RL algorithm}
We apply Distributional Maximum a-posteriori Policy Optimization (DMPO), a variant of Maximum a-posteriori Policy Optimization (MPO)~\cite{abdolmaleki2018maximum} with a distributional critic on the residual task, as it outperformed pure MPO and another recent off-policy algorithm, D4PG~\cite{barth2018distributed}, in our experiments. The implementation is publicly available as part of the Acme framework~\cite{hoffman2020acme}.
MPO is an off-policy actor-critic algorithm based on maximum entropy RL. It treats RL as an inference problem: at each update step, a distribution of policy parameters is updated using a single-step temporal difference (TD) update, with a Gaussian prior around the current policy.
Distributional MPO (DMPO) additionally adopts a distributional Q-network parametrized as in C51~\cite{bellemare2017distributional}: instead of predicting the expectation of $G^\pi(s, a)$, the critic learns to model its full distribution with a categorical distribution of 51 atoms.

\subsection{Training details}
For data efficiency, we use image augmentation in BC training as presented by~\cite{strudel2019learning}, namely random crops and rotations. For Mime tasks, we additionally use the environments' built-in viewpoint augmentation and record the demonstrations from five camera positions which are sampled from a section of a sphere centered on the robot. For details, see \cite{strudel2019learning}. Image inputs are normalized to be in $[-1, 1]$ and the demonstrated actions as well as proprioceptive features are normalized per dimension to have zero mean and unit variance in demonstration data. For tasks requiring the gripper, we empirically set $\lambda$ to $0.9$ as done by \cite{strudel2019learning}.

For $f_\theta$, we use a ResNet-18~\cite{he2016deep} with all layer sizes halved, such that the bottleneck layer features have dimensionality $d=256$, and concatenate the scalar inputs, where used, to the bottleneck features. Given demonstration datasets of different sizes, we use 95\% of the data for training and validate on the remaining 5\%, with the $f_\theta$ training epoch with the lowest validation loss evaluated in the environment on a fixed set of 100 unseen initial states. To represent the residual policy $\pi_\phi^r$, we use a three-layer fully connected neural network with layer normalization and layer sizes $[256, 256, 256]$, and another with sizes $[512, 512, 512]$ for the critic (defaults of the DMPO implementation \cite{hoffman2020acme}).

Given the normalization scheme applied to demonstrated actions in BC training, actions drawn from $\pi_\theta^b$ must again be \emph{denormalized} to match the original action space of the environment:
\begin{align}
u&=\sigma(A) \circ \pi_\theta^b(s)+\mu(A),
\end{align}
where $A$ is the set of all actions in the BC dataset and $\circ$ denotes element-wise multiplication. Like base policy actions, we also denormalize residual actions sampled from $\pi_\phi^r$ using statistics from the demonstrations, but with the output distribution centered around zero rather than the demonstrations' mean:
\begin{align}
a &\sim c \sigma(A) \circ \mathcal{N}(\pi_\phi^r(s, \pi_\theta^b(s))),
\end{align}
where $c$ is an optional scalar. This allows both $\pi_\theta^b$ and $\pi_\phi^r$ to better handle a potentially very heterogeneous action space and efficiently learn tasks that may require significantly different distributions of actions per action dimension. The residual denormalized actions may additionally be scaled using $c<1$ to encourage small corrective actions relative to the variance of demonstrator actions. In addition to $c$, we find the size of exploration (the diagonal elements of $\Sigma$) at initialization to be an important hyperparameter: a larger $\Sigma$ means larger exploratory residual actions will be taken early on in training. We set both $c$ and $\Sigma$'s diagonal (a single value for all elements) by grid search over $\{0.01, 0.033, 0.1, 0.33, 1\}$.
The range for the critic's value distribution is set to $[0, 1]$ to match the range of possible sparse returns, and the learning rate is set empirically to $3.3 \times 10^{-4}$ for all tasks (by considering a small grid around the implementation default, $10^{-4}$). All other algorithm hyperparameters were kept at default values of the Acme implementation.

\begin{figure*}[t!]
    \centering
    \includegraphics[width=\linewidth]{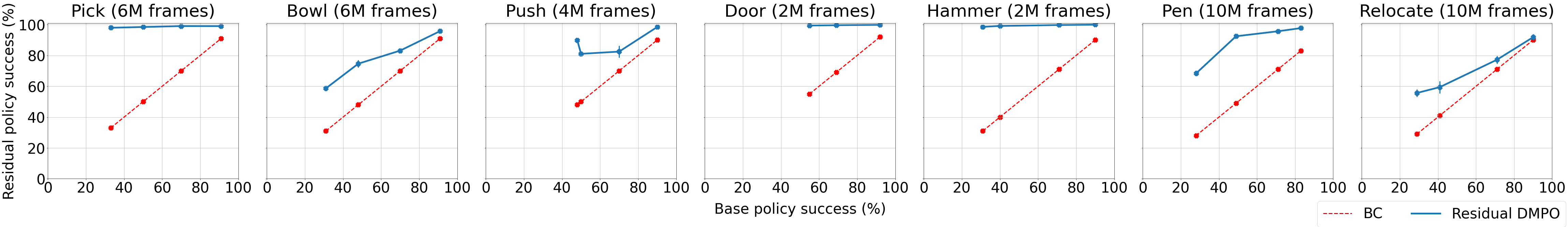}
    \caption{Success rates of residual policies as a function of base policy success rate
    (mean of 5 seeds, 95\% confidence intervals).}
    \label{fig:final_residual_results}
\end{figure*}

\begin{figure*}[t!]
\begin{subfigure}{0.339\textwidth}
\includegraphics[width=\linewidth]{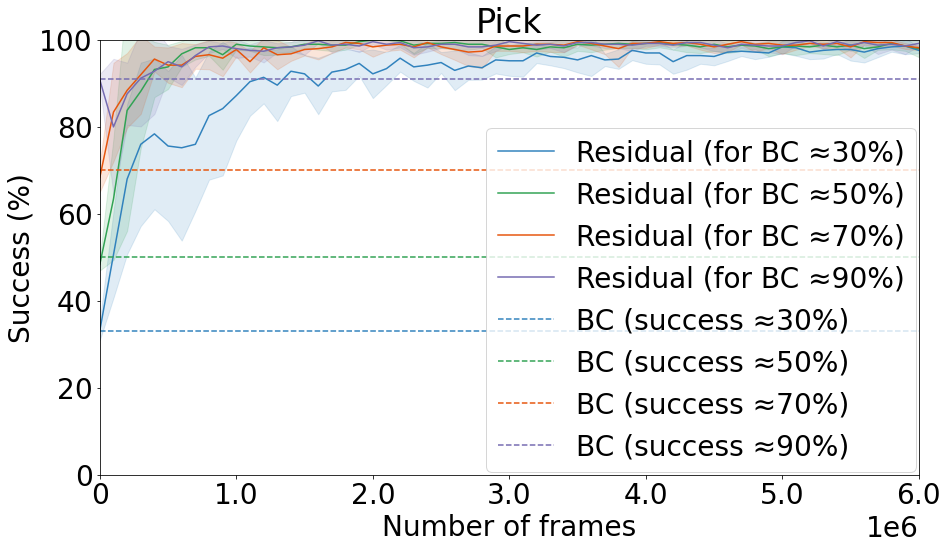}
\end{subfigure}
\centering
\begin{subfigure}{0.324\textwidth}
\includegraphics[width=\linewidth]{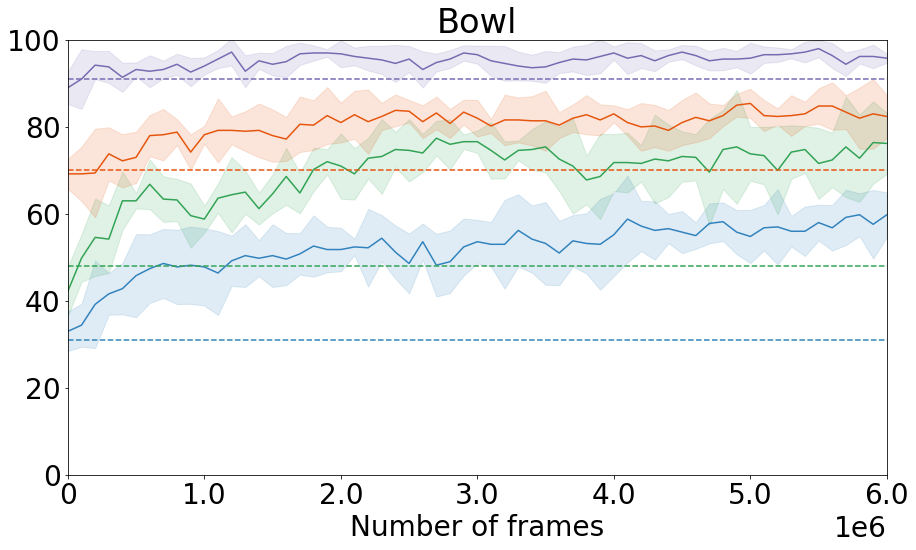}
\end{subfigure}
\begin{subfigure}{0.324\textwidth}
\includegraphics[width=\linewidth]{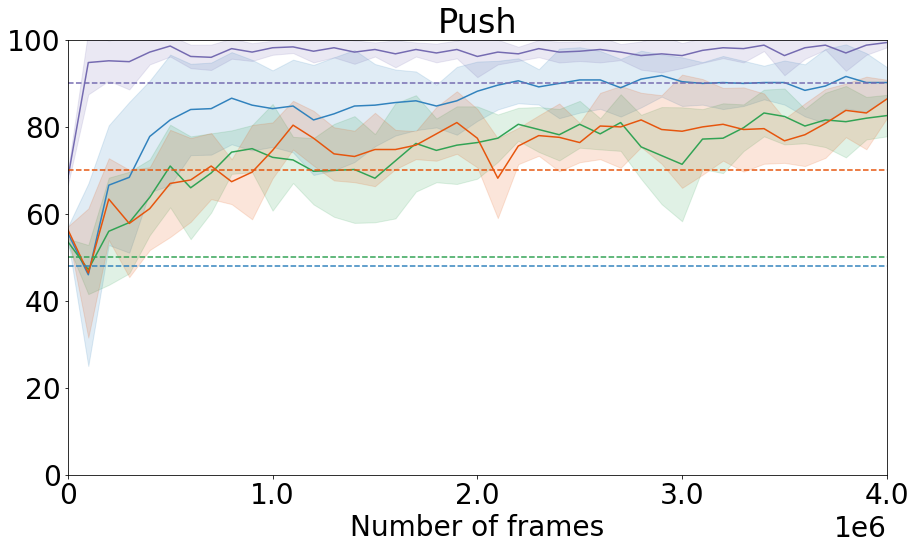}
\end{subfigure}\\
\begin{subfigure}{0.255\textwidth}
\includegraphics[width=\linewidth]{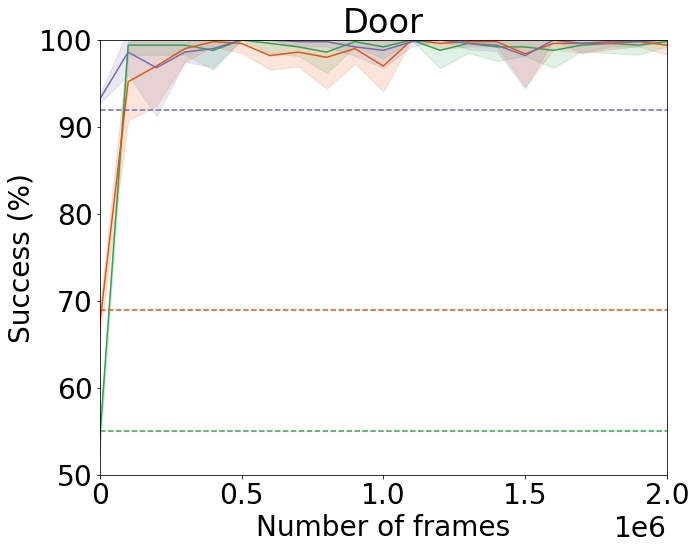}
\end{subfigure}
\begin{subfigure}{0.242\textwidth}
\includegraphics[width=\linewidth]{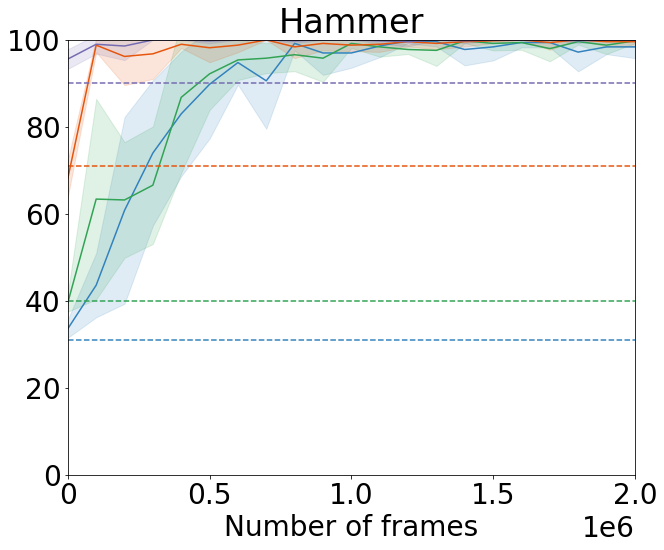}
\end{subfigure}
\begin{subfigure}{0.242\textwidth}
\includegraphics[width=\linewidth]{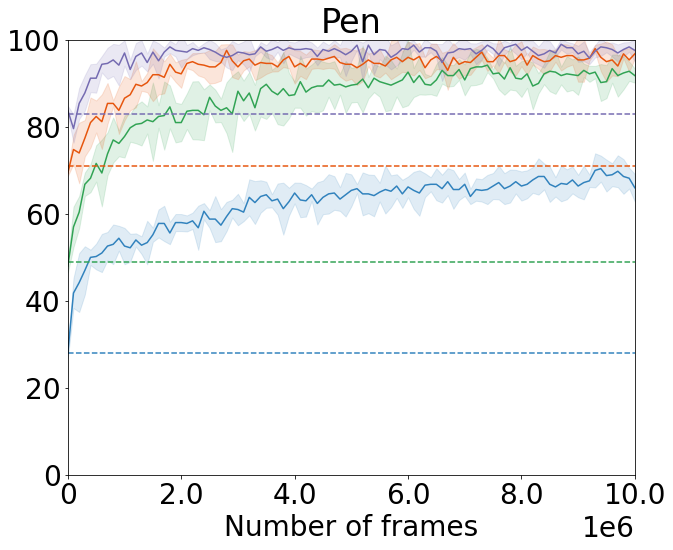}
\end{subfigure}
\begin{subfigure}{0.242\textwidth}
\includegraphics[width=\linewidth]{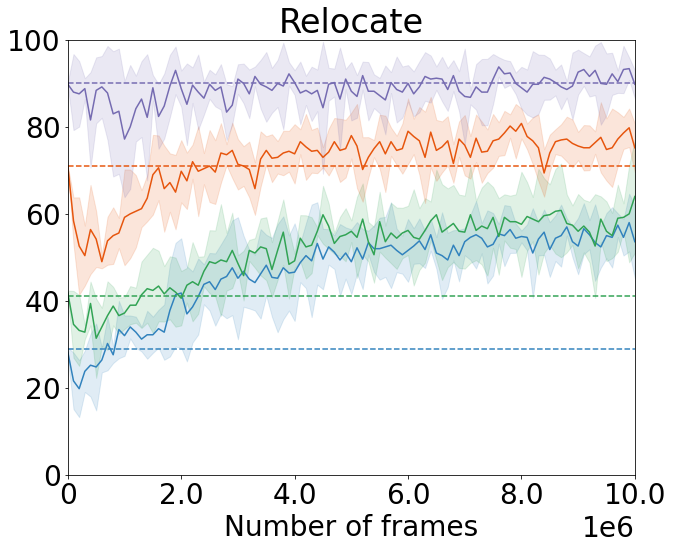}
\end{subfigure}
\caption{Success rates for the residual agent over training (5 seeds, 95\% confidence intervals).}
\label{fig:residual_results}
\centering
    \vspace{0.2cm}
    \includegraphics[width=0.339\linewidth]{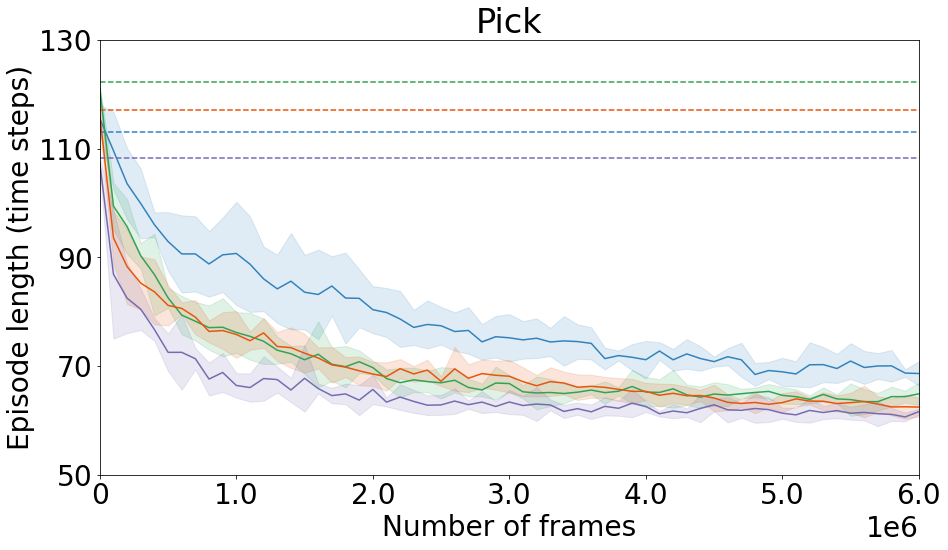}
    \includegraphics[width=0.324\linewidth]{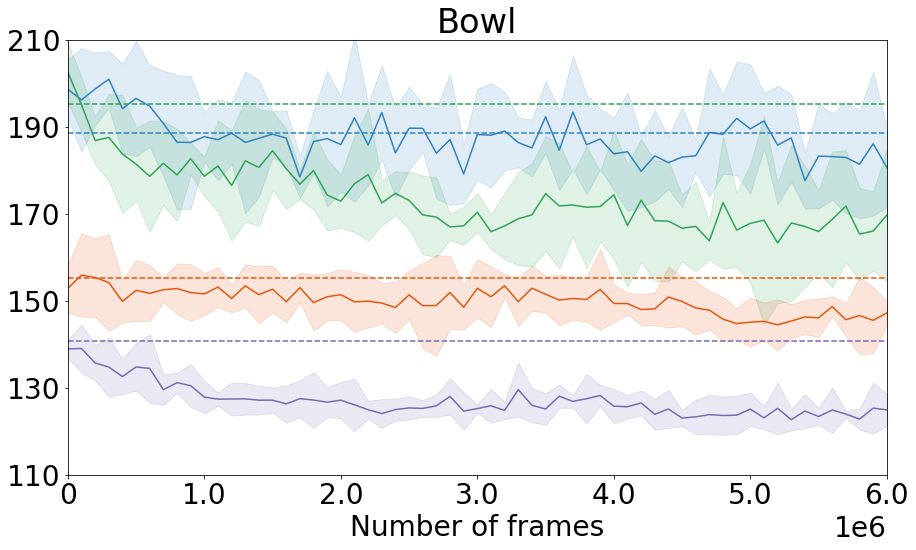}
    \includegraphics[width=0.324\linewidth]{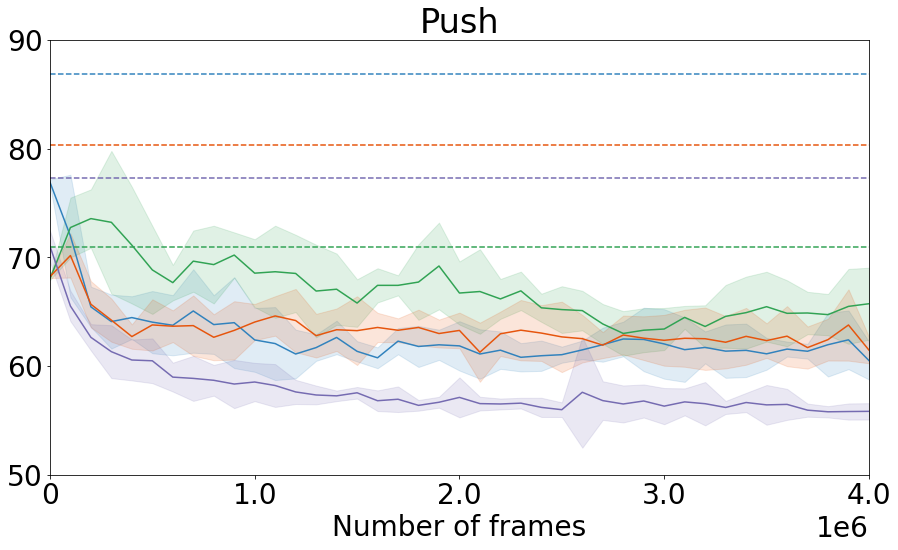}\\
    \includegraphics[width=0.255\linewidth]{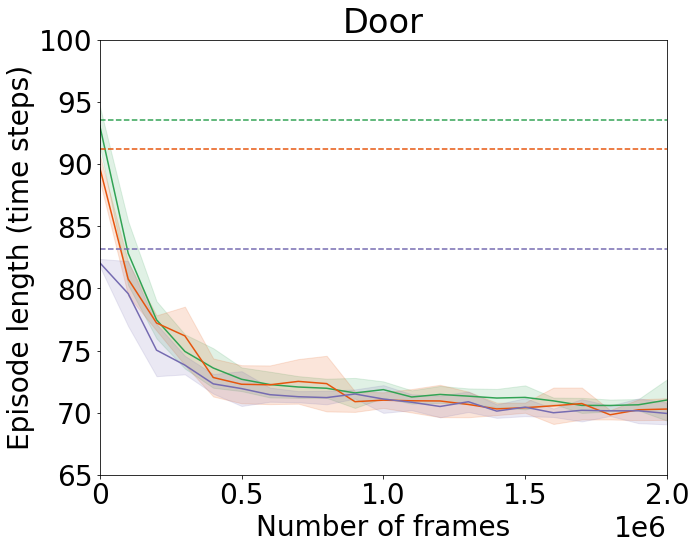}
    \includegraphics[width=0.242\linewidth]{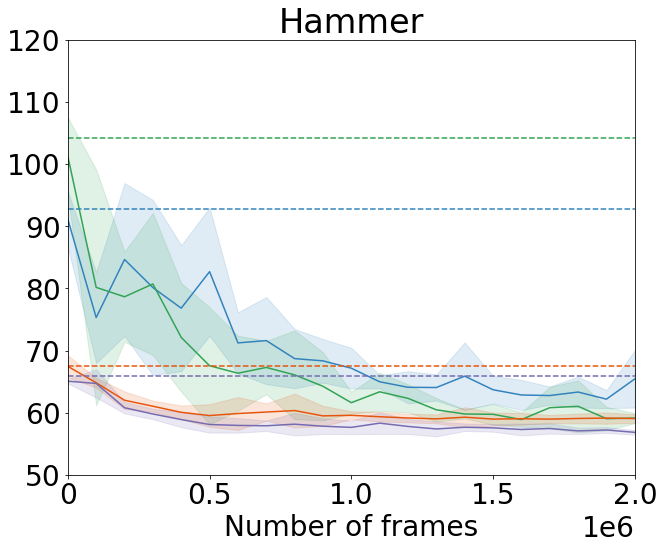}
    \includegraphics[width=0.242\linewidth]{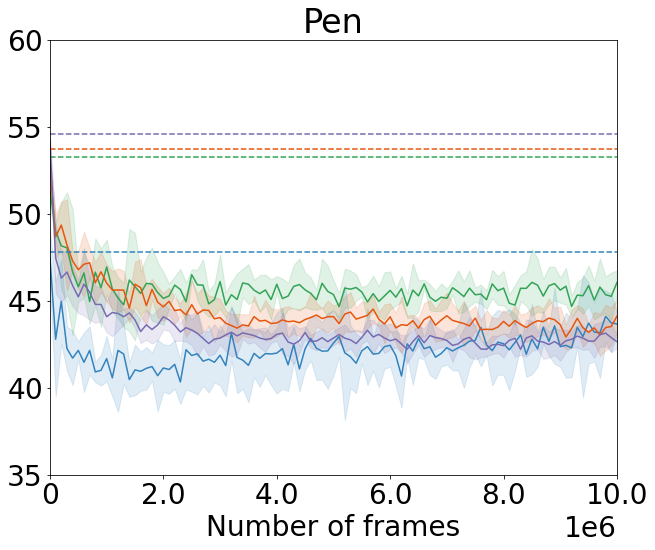}
    \includegraphics[width=0.242\linewidth]{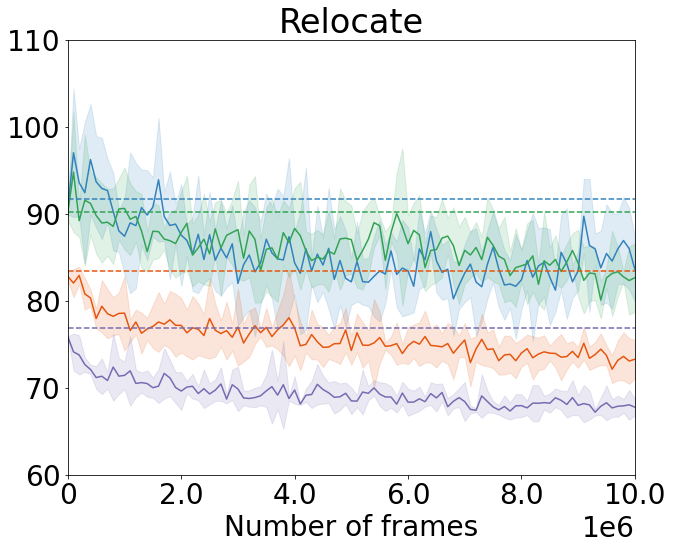}
    \caption{Length (in number of time steps) of successful episodes for the residual agent over training (5 seeds, 95\% confidence intervals). A lower number means the policy is able to solve the task faster.}
    \label{fig:residual_lengths}
\end{figure*}

\subsection{Base policy}
Success rates for the BC policy are given in Fig. \ref{fig:bc_results}. Although there is significant variance across random seeds for most tasks, overall the performance increases as more demonstrations are added and eventually approaches the expert's performance. We observe that conditioning BC on the robot proprioceptive state in addition to images improves performance on Mime but not on Adroit. We believe this is due to 1) the tasks being solvable from images alone, and 2) overfitting enabled by the high dimensionality of proprioceptive state in Adroit (54--100, depending on the task). 
Exploring architectures and training procedures that would enable successful fusion of these modalities, leading to a performance increase on Adroit relative to image-only inputs, would be an interesting topic for further work.

A further increase in success rate can be obtained when full environment state is included; however, we assume it is not known in general. We therefore only consider base policies without access to full state in the residual experiments.

\subsection{Residual policy}
We train residual DMPO on top of BC policies of various strengths to investigate the effect of base success rate on final performance and the required training time. Out of the BC policies included in Fig. \ref{fig:bc_results}, we choose policies with success rates close to 30\%, 50\%, 70\% and 90\% for each task (with the exception of Door, for which no BC policy achieved less than 55\%) to cover a wide range of possible base policy performances. The residual success rate at convergence 
is shown in Fig. \ref{fig:final_residual_results} as a function of base success. 
However, significantly shorter training times are sufficient to outperform the base policies: Fig. \ref{fig:residual_results} shows the improvement in success rate over training. Notably, we observe only a minor drop, if any, in success rate relative to the BC policy at the start of training, making residual training viable on real-world systems where performance should not drastically fall. The policies are evaluated every 100~000 training frames (i.e. steps in the environment); in reporting final numbers, we consider a moving average of 5 evaluations.
 
Although Adroit environments are originally defined with full state observations and shaped rewards \cite{rajeswaran2018learning}, we make use of neither and instead use camera observations, proprioceptive state (joint positions, joint velocities, palm position as well as tactile sensor readings -- the observation space previously defined in \cite{jain2019learning}), and sparse rewards only. Accordingly, we measure success rate of the resulting policies in our evaluation, not episodic return as defined by the shaped reward functions. 
We observed that the RL policy, unlike BC, is able to successfully combine robot state information with visual inputs in Mime as well as in Adroit, and omitting robot state consistently hurt residual performance.

\begin{figure}[t!]
    \centering
    \begin{subfigure}{\linewidth}
    \centering
    \includegraphics[width=0.239\linewidth]{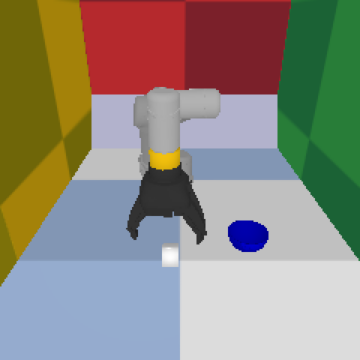}
    \includegraphics[width=0.239\linewidth]{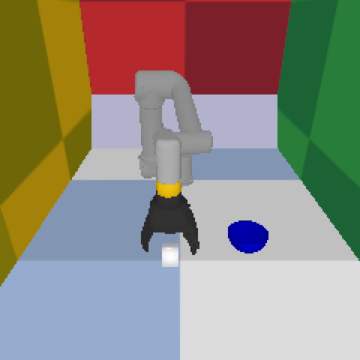}
    \includegraphics[width=0.239\linewidth]{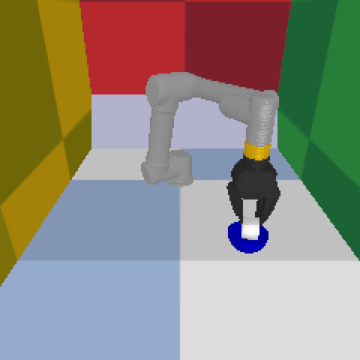}
    \includegraphics[width=0.239\linewidth]{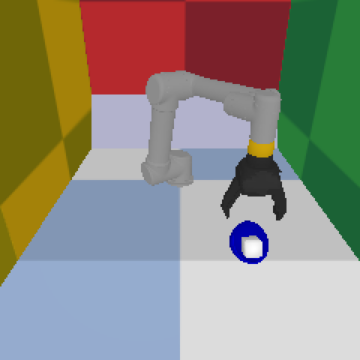}
    \vspace{-1mm}
    \end{subfigure}\\
    \begin{subfigure}{\linewidth}
    \centering
    \includegraphics[width=0.239\linewidth]{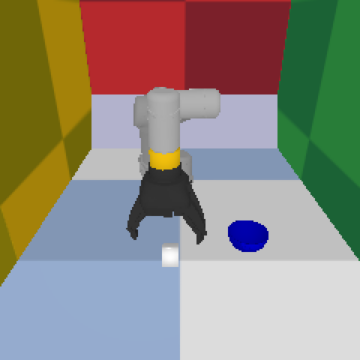}
    \includegraphics[width=0.239\linewidth]{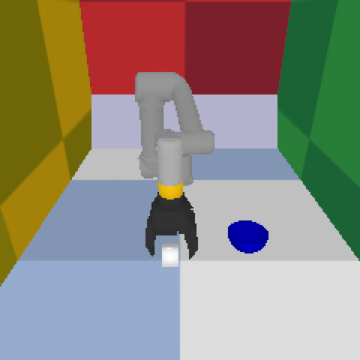}
    \includegraphics[width=0.239\linewidth]{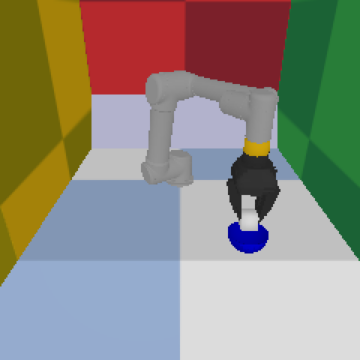}
    \includegraphics[width=0.239\linewidth]{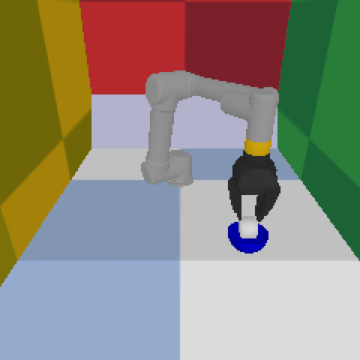}
    \end{subfigure}
    \caption{An example episode from the test set for Bowl where BC fails but the residual policy succeeds. The instance is difficult as the cube is large relative to the bowl. Residual RL has learned to adjust the position from which the cube is released to be more precisely centered on the bowl. Top: BC policy trained on 30 demonstrations. Bottom: Residual policy.}
    \label{fig:bowl_rollouts}
\end{figure}

The average length of successful episodes over the course of training in shown in Fig.~\ref{fig:residual_lengths} for each task. The residual policies learn to solve the tasks not only more reliably, but also more efficiently than BC for all tasks and base policies while respecting the environments' velocity limits, with an average 21\% reduction in execution time, corresponding to 2 seconds of robot time per episode. An example side-by-side comparison of BC and the residual policy is shown in Fig.~\ref{fig:bowl_rollouts}.

\subsection{Fine-tuning baseline}
RRLfD's residual shape with a fixed base policy protects against catastrophic forgetting, which can be a significant source of instability in deep RL. One alternative way to incorporate demonstrations as a prior in RL training is to initialize the policy network using an imitation learning method, such as BC, and then continue training using a non-residual RL objective. We considered two settings: we either froze the ResNet weights as in residual training, or included them in fine-tuning. The network architecture (as shown in Fig. \ref{fig:overview_right}) is kept constant, except for the omission of the base action $u_t$; the network $\pi_\phi^r$ is included at BC training stage. Moreover, a continuous action space is used, including velocity control of the gripper, as DMPO does not support a discrete-continuous action space. Appropriately handling a hybrid action space could be considered in further work, as was done by \cite{neunert2020continuous}. 

We found that fine-tuning the full network leads to immediate catastrophic interference in the ResNet features given that they are shared between the actor and the critic. As the critic network cannot be directly initialized from demonstrations without rewards, updating the full network using the randomly initialized critic's objective destroys the pretrained features -- we found this to be the case even for very small learning rates. As the pretrained policy in turn depends on these features, its performance falls to near 0\%, after which no further useful training data can be collected.

In an attempt to alleviate catastrophic forgetting, we also run experiments with the CNN weights fixed and fine-tune the policy layers $\pi_\phi^r$ only. Although the fine-tuned policy is able to improve upon BC for some of the easier tasks, it requires longer training and larger exploratory actions than residual RL, both of which would be undesirable on a real robot. For Door, success rates of 97.9\% and 99.4\% rates are achieved given initializations with 75\% and 89\% success, respectively, after 3 million frames and only with the diagonal elements of $\Sigma$ initialized to $1.0$ ($0.33$ is sufficient in residual training). 
For Pick and Pen, fine-tuning was able to slightly improve over BC, but only for low-performing initializations: at best, it scored 50.8\% for Pick after 3 million frames, and 52.2\% for Pen after 7 million frames, starting from 30\% in both cases. For all other tasks and BC initializations (policies with $\approx$30 and $\approx$90\% success were considered for each task), success never exceeded BC performance. A direction for further work could be to consider also initializing the value function from demonstrations (see e.g. \cite{wang2020critic}) by labeling demonstrations with a sparse reward, assuming successful completion. However, a perfect demonstrator is not always available: our expert scores just 81\% for Pen, for example.

\subsection{RL baseline}
We have also run experiments to find out the number of frames required to train DMPO from scratch on each task. As input, we consider either images with robot proprioceptive state (denoted robot state) or the full environment state including robot state (denoted full state), in which case the CNN is omitted. Each action dimension is normalized to the unit interval, as demonstration statistics are not assumed to be available.

\begin{table}[t!]
    \centering
    \begin{tabular}{|l|c|c|c|c|c|c|}
        \cline{2-7}
        \multicolumn{1}{c|}{} & \multicolumn{2}{c}{image + robot} &\multicolumn{4}{|c|}{full state}\Tstrut\Bstrut\\
        \cline{2-7}
        \multicolumn{1}{c|}{} & 
        \multicolumn{2}{|c|}{dense reward} &
        \multicolumn{2}{|c|}{sparse reward} & \multicolumn{2}{|c|}{dense reward} \Tstrut\Bstrut\\
        \hline
        task & success & steps & success & steps & success & steps \Tstrut\Bstrut\\
        \hline
        Pick & 0.0 & 8M & 59.2 & 9M & 98.8 & 10M \Tstrut\\ 
        Bowl & 0.0 & 8M & 0.0 & 20M & 52.2 & 9M \\ 
        Push & 1.0 & 3M & 99.0 & 2M & 98.0 & 2M \\ 
        Door & 0.0 & 6M & 20.0 & 5M & 99.5 & 7M \\ 
        Hammer & 6.3 & 6M & 20.0 & 7M & 98.9 & 6M \\ 
        Pen & 9.4 & 4M & 93.1 & 13M & 91.2 & 12M \\ 
        Relocate & 0.0 & 6M & 0.0 & 20M & 97.4 & 12M
        \Bstrut\\
        \hline
    \end{tabular}
    \vspace{0.2cm}
    \caption{Sample efficiency and success rate at convergence of DMPO trained from scratch. Sparse reward policies from images and robot state scored nearly 0.0 for all tasks (not shown), which is not surprising given that RL without any priors is extremely difficult in settings with sparse rewards and very high-dimensional inputs.
    }
    \label{tab:rl_from_scratch_results}
\end{table}

Although our setting of interest is that of sparse task completion rewards and readily available state features, we also consider settings with dense reward signals and full state features for each task to highlight the trade-offs between providing demonstrations and designing shaped rewards or higher-level state features. 
For Pick, we use the negative Euclidean distance between the gripper and the cube; for Bowl, the sum of the negative distance between the gripper and the cube, and between the cube and the bowl; and for Push, the sum of negative distance between the gripper and the cube as well as the cube and the goal region. While further reward shaping beyond our choice of reward functions could accelerate learning, defining a fully dense reward, i.e. one that guides the policy towards solving the task from any state---without creating unintentional local maxima---can in general be as difficult as designing a controller to solve the task, and usually requires balancing multiple reward components, such as approaching the cube vs. bringing the cube closer to the bowl. For Adroit environments, we use the carefully shaped multi-component rewards defined by \cite{rajeswaran2018learning}. We also add sparse task completion to each dense reward, appropriately scaled to prioritize task success, as we found this to increase success rates.

As shown in Table~\ref{tab:rl_from_scratch_results}, learning from images does not lead to any successful policies within the training budgets:
although the agent occasionally completes an episode successfully, this is not sufficient to learn useful visual representations, even with a dense reward. It is difficult and time-consuming to learn control from scratch without any priors in this setting due to the high dimensionality of both the input and the action space; many RL tasks consider a smaller dimensionality and use full state based inputs or low-resolution images instead. Replacing the depth frames with the full environment state simplifies the task and regularly leads to successful policies in the sparse reward setting for Push and Pen -- as well as occasionally for Pick, Door and Hammer -- and in the dense reward setting for all tasks. Learning Bowl from scratch is particularly tricky as adept gripper control is of central importance, whereas DMPO considers a continuous action space.

Comparing the empirical sample efficiencies of the RL baselines and the results presented in Fig.~\ref{fig:final_residual_results}, we have shown demonstrations to be a viable alternative to manual reward shaping and full state estimation. Using a residual formulation, learning from demonstration can be combined with any RL method.


\section{Discussion} 
\label{sec:conclusion}
We have presented a novel method for combinining demonstrations and RL to learn continuous control through a residual formulation. Our approach does not require base controller engineering, reward shaping or feature engineering as it is able to learn from readily available image and proprioceptive inputs and sparse rewards. We have empirically demonstrated sample-efficient training on robotic manipulation tasks in simulation, in a setting out of reach for RL from scratch with current algorithms. What is obviously missing in our presentation are results on a physical robot.
The current pandemic conditions have prevented us from conducting such experiments for this submission. We will evaluate our method in this setting after the situation improves.

As the base policy is treated as a black box during residual training, the proposed method is equally compatible with hand-engineered base controllers as with the BC policy considered in this work.
Provided a sufficient quantity of demonstration data is available, using a BC policy as the base controller represents a more flexible choice that may be improved as more data becomes available. Data gathered on the system, which BC directly models, better represents true environment dynamics than is possible with first-order modelling in hand-engineered controllers without extensive manual tuning, especially in tasks with rich contacts and friction and in high-dimensional action spaces. In a low-data regime, however, and when accurate state estimation is feasible, using a classical controller remains a good alternative. In this setting as in the case of BC base policies, a residual in image space is a valuable addition as it can learn to correct for state estimation noise or omission of relevant state features. Task-specific visual features learned through BC on collected trajectories can equally be used in the classical residual RL setting to accelerate learning from images on the residual control task.

\addtolength{\textheight}{0cm}   





\section*{Acknowledgment}

This work was supported in part by the Inria / NYU collaboration, the Louis Vuitton / ENS chair on artificial intelligence and the French government under management of Agence Nationale de la Recherche as part of the “Investissements d’avenir” program, reference ANR19-P3IA-0001 (PRAIRIE 3IA Institute). M. Alakuijala was supported in part by a Google CIFRE PhD Fellowship. J. Mairal was supported by the ERC grant number 714381 (SOLARIS project) and by ANR 3IA MIAI@Grenoble Alpes, (ANR-19-P3IA-0003).


\bibliographystyle{IEEEtran}
\bibliography{IEEEabrv,references}

\end{document}